# Understanding and Predicting The Attractiveness of Human Action Shot

Bin Dai
Institute for Advanced Study,
Tsinghua University, Beijing, China
daib13@mails.tsinghua.edu.cn

Baoyuan Wang
Microsoft Research,
Redmond, US
baoyuanw@microsoft.com

Gang Hua
Microsoft Research,
Redmond, US
ganghua@microsoft.com

## Abstract

*Selecting attractive photos from a human action shot sequence is quite challenging, because of the subjective nature of the "attractiveness", which is mainly a combined factor of human pose in action and the background. Prior works have actively studied high-level image attributes including interestingness, memorability, popularity, and aesthetics. However, none of them has ever studied the "attractiveness" of human action shot. In this paper, we present the first study of the "attractiveness" of human action shots by taking a systematic data-driven approach. Specifically, we create a new action-shot dataset composed of about 8000 high quality action-shot photos. We further conduct rich crowd-sourced human judge studies on Amazon Mechanical Turk(AMT) in terms of global attractiveness of a single photo, and relative attractiveness of a pair of photos. A deep Siamese network with a novel hybrid distribution matching loss was further proposed to fully exploit both types of ratings. Extensive experiments reveal that (1) the property of action shot attractiveness is subjective but predicable (2) our proposed method is both efficient and effective for predicting the attractive human action shots.*

## 1. Introduction

With the ubiquity of camera phones, it is convenient for us to shoot as many photos as desired. However, capturing compelling human action shots remains to be a challenge, even for professional photographers. In order not to miss the best moment shot, photographers typically need to rely on the burst capture mode to shoot several consecutive frames using fast shutter speed, leaving the photo selection as a tedious yet important manual post processing step.

While researchers have proposed various computational methods [5, 4, 31, 27, 15, 11, 7, 23] for automatic photo selection from personal albums, considering common factors such as image technical quality (*i.e.*, blur, noise), visual diversity, memorability [15], interestingness [14], and aesthetic properties [7], little attention has been devoted to the task of *attractive* action shots selection. In the context of action shot photography, the most *attractive* shot within a burst set is largely often determined by the human pose of a specific action, as the burst of photos share the same action context (*i.e.*, the background). For example, in a sequence of Fosbury-Flop, as shown in Figure 1, the most attractive shot should be the frames when jumpers are leaping head first with their back to the bar, which is a brief moment usually called peak-action, as it hovers motionless before starting back down. This raises up an interesting yet challenging question, how can we automate the process for attractive action shots selection?

A direct thought is to analyze the human motions of the image set. However, it is intrinsically challenging to perform the motion alignment and tracking along the shot sequence. In addition, accurate human pose estimation remains a challenge by itself especially for those attractive human poses, despite the fact that dramatic improvement was made in the past two years [1]. More importantly, the human pose is an important factor, but not the *only* factor to determine the most attractive action shot. It is mainly the human body pose combined with the background context ultimately determines if an action shot photo is attractive or not. For example, the same jumping pose of the same person in a kitchen may largely perceived to be less attractive than that in a seashore.

We approach to this problem from another perspective by studying the *general attractiveness* of action shots. We argue that such general attractiveness of action shot photos exists in human perception. For example, most people would perceive that the *Hip-Hop* dance poses are more attractive than normal human walking poses. In skateboarding, the pose at the moment when the skateboarder freezes in the air without touching any objects is more attractive than those poses when he starts to approach from the ground. If we can successfully model the general attractiveness of the action shot, we can then resolve the problem of photo selection in any sequence of action shot, not limited to a burst set, as now the model allows us to assign a global "attractiveness" score to each action shot. Such a score enables direct comparison of the attractiveness of different action shots in different action context. This would further support broader photo selection function in a photo album for applications, such as selecting attractive representative action photos across a personal photo album.

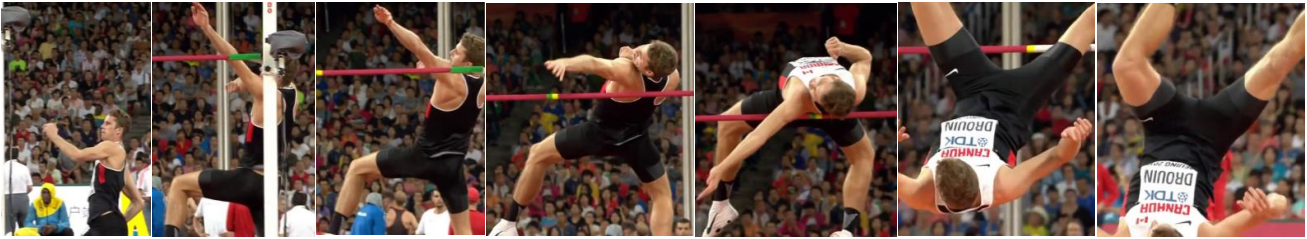

Figure 1. A typical action shot sequence of Fosbury-Flop

However, similar to the concept of attractiveness for portrait photos as discussed in [33], the concept of attractiveness of action shots is also somewhat subjective. Therefore it is very difficult, if not entirely impossible to define the attractiveness of action shot photos in an qualitative way. Hence, we take a data-driven approach by gathering human judge data in terms of both absolute attractiveness rating on individual action shot photos, as well as relative attractiveness rating on pairs of action shot photos, due to the complementary nature of these two rating schemes. Hence, multiple human judges on the same photo or pairs of photos are gathered from Amazon Mechanical Turk (AMT). The details of our action photo collections along with the human judge data can be found in Section 3. Our data collection is consistent with our expectation that there are diverse opinions among the human judges. To deal with the diverse opinions from multiple human judges, many previous works attempt to consolidate the human judge data first [14, 18], *e.g.*, by taking the majority votes, and then learn a model with such consolidated unique ratings. Unlike them, we design a Deep Convolution Neural Network (DCNN) with a hybrid loss function which matches the distribution of the human judge preferences on both the absolute and relative ratings. We argue such a design directly takes the diverse opinions from the human judges into consideration, and hence avoid *ad-hoc*, hand-crafted pre-filtering of the human judge data. Our DCNN based model naturally takes both the human in action and the background context into consideration, and makes it unnecessary to conduct human detection and pose estimation. Our experiments reveal that our learned DCNN model can automatically draw its attention to the human in action and the surrounding background context. To the best of our knowledge, this paper presents the first study on general attractiveness of action shot photos. This work hence presents the following contributions:

- we created a new action shot dataset and collected rich annotations via crowd-sourcing to study the general attractiveness attribute of action shots;

- we designed an efficient hybrid training model based on deep learning to match the rich crowds distributions;

- we demonstrated that the general attractiveness of action shots is subjective but still predictable through our proposed model.

Last but not least, our model can be easily applied and extended in a variety of practical applications including sports video highlight extraction, event curation, and photo album summarizations.

## 2. Related Work

### 2.1. Photo Selection based on High-level Attributes

Automatic photo selection from personal photo collections has been actively studied over the years [5, 4, 31, 27] in both multimedia and computer vision. The selection criteria, however, is primarily focused on low-level technical image quality, representativeness, diversity and also coverage. Recently, there has been as increasing interest in understanding and learning the various high-level image attributes, including memorability [15, 16, 21, 10], popularity [20], interestingness [11, 14, 7, 8], aesthetics [7, 22, 6, 8], importance [2] and specificity [17]. Extensive studies were also conducted to uncover the relationships among these attributes. For example, It is found that there exists a strong correlation between interestingness and aesthetics while surprisingly no correlation exists between interestingness and memorability[14]. Although these prior works are relevant, our work is distinct in a number of ways: (1) We focus on people-centric images with specific to action shots, while prior works consider more generic scene categories such as landscapes. (2) We are interested in studying the main factors of human body pose and its context (i.e. background) in determining the attractiveness of an action shot, assuming other factors such as technical quality are the same. Therefore, any discriminative features such as blurriness, rule of thirds for measuring atheistic attribute may not apply for determining the attractive human action shots. (3) Due to the problem differences, there are no existing datasets can be directly used to train an attractive action shot detector. Therefore, we created our own datasets and used AMT to tag each action shot image primarily based on the human pose attractiveness for specific actions. To the best of our knowledge, no prior work has ever put human pose into consideration for studying any of these high-level attributes. (4) Most of prior works heavily reply on hand crafted features to learn the discriminative models, We instead leverage the recent advances of deep learning and try to build the high-level representations through a novel hybrid distribution matching loss.

## 2.2. People-Centric Image Understanding

Understanding people-centric images is always the most important yet challenging problem in computer vision. Over the past years, dramatic improvement has been made in the subproblems from people/pedestrian detection [9], pose estimation [1] and single image action recognition [12], to face detection [30], alignment [24] and recognition [25]. Few works have also been devoted to understand the various high-level attributes such as recognizing the different facial expressions [29, 19] and predicating the attractiveness of a portrait image [33]. However, still no attention has been payed for studying the attractiveness of a human action shot, which essentially has plenty of important applications in computational photography and multimedia.

## 3. Dataset

We collected an action-shot image dataset crawled from the Internet using Google image search. We search images on Google using both general keywords such as "action shot", as well as keywords covers various sports domains including *soccer*, *basketball*, *tennis*, *surfing*, *skating*, *skiing*, *dancing*, and *gymnastics*, *etc*. We also used the names of famous sport stars and specific actions to expand the coverage and diversity of our collected action shot photos. For example, *Messi shot*, *Kobe layup*, *Iverson dribble*, and *skateboarding tricks etc*. Over 12,000 images were collected from Google image search initially. In order to study the main factors of human body pose and the background context, we further remove the low-resolution and low quality (such as blur, noisy *etc*) ones, which leads to 7980 valid images in our final collection. In order to study the attractiveness of these action shot photos, we use AMT to evaluate both the absolute attractiveness ratings on single images, and relative attractiveness ratings on image pairs, to leverage the complementary nature of these two different rating schemes.

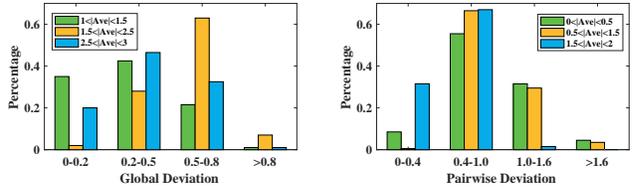

Figure 2. Rating Distributions. *left:* The distribution of global ratings. It is obvious that the samples with average scores around 2 have large deviations. *right:* The distribution of pairwise ratings. The pairs with absolute average vales close to 2 have smaller deviations. The attractiveness of the two samples in such pairs greatly differ with each other.

### 3.1. Global attractiveness rating

We exploit AMT to rate each image with 1 to 3 stars. An image rated with 1 star means it is not an attractive action shot. For example, a person with an standing posture is certainly not an action shot. In contrast, a 3-star rating means that the image is definitely an attractive action shot. For example, a person flipping in the sky is absolutely an action shot. However, there are cases which are quite difficult to tell. For example, in a soccer match, could dribbling be regarded as an attractive action shot? The pose is quite representative and could not be persistent even for a second. However, it is not that attractive. For these cases, one may rate it with 2 stars.

We asked 10 people to rate each image with 1 to 3 stars. Thus we obtain a probability distribution on the three ratings of each image. Although the rating is quite subjective, we can still find some consensus across different people. As shown in Figure 2 *left*, more than 30% samples whose average rating is less than 1.5 are with deviation less than 0.2. People generally have similar preferences on the most (blue bars) and the least (green bars) attractive images. For the images having nearly 2 average stars (yellow bars), people tend to have different preferences.

This manifests that there are some consensus among all the people on if an image is an attractive action shot or not. However, our global ratings, especially the middle part, is subject to large variations, as we expected.

### 3.2. Pairwise ratings

Though the global ratings roughly indicate to us how likely an image is to be an attractive action shot, it is still quite noisy. Such a rating method could not catch the subtle differences between images. To obtain more subtle information, we use pairwise labeling. At each time, two images are presented and the Turkers are asked to rate the relative attractiveness between the two images at 5 levels, *i.e.*, the first image is *much better than{2}/slightly better than{1}/equally good to{0}/slightly worse than{-1}/much*

|         | Better | Equal | Worse |
|---------|--------|-------|-------|
| Better  | 0.94   | 0.03  | 0.04  |
| Equal   | 0.24   | 0.56  | 0.20  |
| Worse   | 0.10   | 0.31  | 0.59  |

Table 1. Confusion Matrix For $c_g = 0.3$ and $c_p = 0.2$

*worse than*{-2} the second image.

Since there are $N^2$ pairs if we have $N$ images in total, it is very difficult, if not impossible, to rate every pair in our dataset. Randomly sampling pairs would be a choice. However, we wish to sample more pairs which share similar appearances. So we first extract the appearance feature (fc7 layer output of the VGG16 [26]) of each image and then conduct $L2$ normalization on the features. Denote all the images as $I_1, I_2, ..., I_N$ and their features as $f_1, f_2, ..., f_N$. Note that $||f_i||_2 = 1, i = 1, 2, ..., N$. For each image $I_i$, we randomly sample 5 pairs from the rest images. The probability for $I_j (j \neq i)$ to be selected is defined as

$$p_j = \frac{exp(f_i \cdot f_j)}{\sum_{j' \neq i} exp(f_i \cdot f_{j'})} \quad (1)$$

In this way, we produced $5N$ pairs which is much less than $N^2$. Each image appears in 10 pairs on average. For each pair, we asked 5 people to rate the relative attractiveness of the two photos. Again we obtain the probability distribution over the 5 different relative ratings.

The pairwise labeling distribution is shown in Figure 2 *right*. In the case that the two images have a large gap in terms of attractiveness, the deviation is quite small (blue bars). This is sensible. If the two images are equally attractive on average (green bars), it either means they are actually equally good (low deviation cases) or people have different opinions on the two images (high deviation cases).

Note that the average value of relative attractiveness ranges from $-2$ to $2$ while that of the global attractiveness ranges from 1 to 3. The deviations of the two cases also have a ratio of 2. Actually the mean deviation of global case is $0.457$. The mean deviation of pairwise case is $0.776$, which is smaller than twice the value of the global case. We can conclude that the pairwise rating has relative low deviation. It indicates that people have more consensus when rating the relative attractiveness of image pairs.

### 3.3. Comparing the two types of ratings

We anticipate that these two types of rating methods agree with each other in general but complement each other in some specific details. To verify that, we analyze the ratings in the following way. We calculate the average global rating $ave_g$, $(1 \leq ave_g \leq 3)$, for each image, and the average deviation $ave_p$, $(-2 \leq ave_p \leq 2)$, of relative rating for each image pair. For the global rating, we regard the two images as equally attractive if $|ave_{g,1} - ave_{g,2}| \leq c_g$, where $c_g$ is a threshold. Otherwise the image with a higher average score is considered to be more attractive. As to the pairwise ratings, we regard the two images as equally attractive if $|ave_p| \leq c_p$, where $c_p$ is another threshold. Otherwise the first image is more attractive if $ave_p < 0$ and the second image is better if $ave_p > 0$.

By setting the two thresholds $c_g$ and $c_p$ at different levels, we find that about $60\% - 75\%$ pairs agree with each other under these two rating methods. The percentage changes with different thresholds. A closer look reveals that most of the disagreements are where one method gives an equal rating while the other gives a more attractive or less attractive rating. To demonstrate this, let us study the case where $c_g = 0.3$ and $c_p = 0.2$. This would result in a $70\%$ agreement. The confusion matrix under that setting is presented in Table 1. The percentage that the two different ratings return opposite ratings is very small. This fact implies that there are some certain consensus perception of the attractiveness of an action shot, and hence make the modeling possible.

## 4. The Deep Siamese Network with Hybrid Distribution Matching Loss

An overview of the network structure of our proposed DCNN model with hybrid distribution matching loss is presented in Figure 3, which exploits a Siamese structure and models both the global ratings and the pairwise ratings. As shown in Figure 3, the foundational unit of the proposed model is the score net. The score net adopts layers from conv1 to conv5 (including pool5) of the VGG16 network [26], which outputs a 512-channel $7 \times 7$ feature map.

On top of it we add two fully convolutional layers along with a ReLU layer, which leads to a 128-channel $7 \times 7$ feature map. Then we conduct a spatial max pooling to get the 128-dimension feature of the image, which is subsequently fed into a fully connected layer to compute a single score $s$ of the image. The higher the score $s$ is, the more attractive the image is as an action shot. The input to the score net is a normalized image of size $224 \times 224$.

Since each individual image appears in five pairs of images in our relative ratings, we use a pair of image as one training sample. In the training process, we consider the combined loss function from both the global ratings and the pairwise ratings, where we introduce a hybrid loss function that matches the distributions of both global and pairwise ratings from crowds. Denote a training image pair as $\mathbf{I} = \{I^1, I^2\}$ and the scores of the pair (after running through the same score net) as $\mathbf{s} = \{s^1, s^2\}$. Let the number of global attractiveness ratings be $M_g (= 3)$, and the number of relative attractiveness ratings be $M_r (= 5)$. It should be noticed that $M_r$ must be odd since the relative attractiveness label is symmetric. For this reason, we define another number $R = (M_r - 1)/2$.

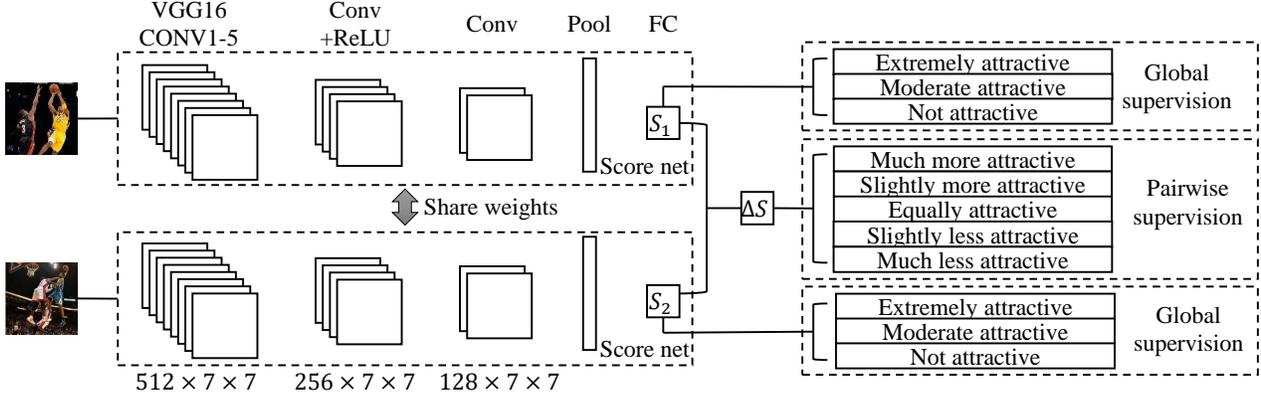

Figure 3. Our Siamese model with hybrid distribution matching loss.

To match with the global ratings, we introduce a set of parameters $\hat{s}_{g,i}$, $i = 1, \ldots, M_g$, namely the standard scores, which are all learned by back propagation through the network. The probability of the $i$-th rating for the $j$-th ($j = 1, 2$) image in the training pair could then be expressed as

$$p_{g,i}^j = \frac{exp(-(s^j - \hat{s}_{g,i})^2)}{Z_g^j} \quad (2)$$

where $Z_g^j$ is the normalization factor and defined as

$$Z_g^j = \sum_{i=1}^{M_g} exp(-(s^j - \hat{s}_{g,i})^2) \quad j = 1, 2. \quad (3)$$

Note that we do not view this problem as a classification problem because we do not even have a unique "ground-truth" rating for each image. So we just train our network to make its predicted distribution $p_{g,i}^j$ to match with the distribution $\hat{p}_{g,i}$ of the global ratings from crowds, where $\hat{p}_{g,i}$ can easily be computed from the set of global ratings for each image. To achieve this goal, we adopt the cross entropy loss which is defined as

$$L_g^j = \sum_{i=1}^{M_g} \hat{p}_{g,i}^j log(p_{g,i}^j) \quad (4)$$

To match with the pairwise ratings, we calculate the gap between the two scores $\triangle s = s^1 - s^2$. Similarly, we could obtain the distribution on the $M_r$ ratings according to a set of parameters representing the standard relative rating scores, $\triangle \hat{s}_{r,i}$, $i = -R, \ldots, R$. Note here only $R$ standard relative rating scores are needed due to the symmetry property of relative ratings. Specifically, we define $\triangle \hat{s}_{r,1}(> 0)$ for slightly more attractive and $\triangle \hat{s}_{r,2}(> \triangle \hat{s}_{r,1})$ for much more attractive. The standard score $\triangle \hat{s}_{r,0}$ is 0 for equally attractive case, then $\triangle \hat{s}_{r,-1} = -\triangle \hat{s}_{r,1}$ would represent the standard score for slightly less attractive and $\triangle \hat{s}_{r,-2} = -\triangle \hat{s}_{r,2}$ for much less attractive. These parameters can also be learned by back propagation through the network. Similar to the case of global ratings, the probability of the $i$-th relative attractiveness, $i = -R, \ldots, R$, is then defined as

$$p_{r,i} = \frac{exp(-(\triangle s - \triangle \hat{s}_{r,i})^2)}{Z_r} \quad (5)$$

where $Z_r$ is the normalization factor and defined as

$$Z_r = \sum_{i=-R}^{R} exp(-(\triangle s - \triangle \hat{s}_{r,i})^2) \quad (6)$$

Again, we adopt the cross entropy loss to match the crowds ratings of the relative attractiveness on the image pair. The loss function is expressed as

$$L_r = \sum_{i=-R}^{R} \hat{p}_{r,i} log(p_{r,i}), \quad (7)$$

where $\hat{p}_{r,i}$ is the distribution of the relative ratings from crowds on the image pair. It can be easily computed by counting the number of relative ratings falling into each buckets.

As we can imagine that supervision from the global ratings cannot catch the subtle differences between pairs of images, while the supervision from the pairwise ratings is lack of global reference, so they complement with each other. To best leverage their complementary power, we argue that their relative importance for the training should be adaptive to each pair. Hence, we define the final loss function for each training pair as:

$$L = \lambda \cdot (L_g^1 + L_g^2) + (1 - \lambda) \cdot L_r, \quad (8)$$

where $\lambda$ denotes the adaptive weight of the two type of loses. Intuitively, if the two images from a training pair have similar global distribution, the pairwise supervision is more important than the global information. In this case, the global supervision may even be misleading since it forces the two images to have similar scores rather than separate them apart. However, if the global distributions of the two images are very different, the global supervision should be dominant, because the pairwise supervision may be redundant in this case. Therefore, we adaptively define the weights of the two kinds of supervision for each training pair according to the similarity of the distribution of the global ratings, *i.e.*,

$$\lambda = ||\hat{p}_g^1 - \hat{p}_g^2||_2^2 \quad (9)$$

where $\hat{p}_g^j = (\hat{p}_{g,1}^j, ..., \hat{p}_{g,M_g}^j), j = 1, 2$. Conceptually, global supervision would coarsely tune the network to ensure the rough attractive order while pairwise supervision fine tune the network to learn their local and subtle relative orders. So they dominate in different cases and are well complement each other. If we have $T$ training image pairs, denote $L_t$ as the loss for the $t$-th training pair, then final training loss function is the sum of all the individual losses, $\mathcal{L} = \sum_{t=1}^{T} L_t$. All the parameters of the network, along with the standard rating scores introduced in the loss function, are optimized via back propagation using stochastic gradient descent.

## 5. Experiments

We randomly selected 5980 images in the collection as training data and left the rest 2000 images as testing data. The pairs were sampled within training data or testing data without overlapping. In other words, there is no pair with one image in the training data and the other in the testing data. In the first stage, we fix the VGG16 [26] part and only train the new layers for 2 epochs with the learning rate $\lambda = 1e-6$. In the second stage, we loose the previous VGG layers and train another 6 epochs. The initial learning rate in this stage is set to $\lambda = 1e-6$ and it scales down 10 times every 2 epochs.

### 5.1. Evaluation metrics

Before we could evaluate our experimental results, we need define some meaningful evaluation metrics as our learning objective is to match the distribution of crowds ratings. One direct measure would be to evaluate how well the predictive distribution matches with the crowds ratings in the test data. However, this metric by itself is not straightforward to understand. Hence in our evaluation, we adopted some more direct evaluation metrics. Since we have both global and pairwise ratings, in the following, we derive the two types of evaluation metrics correspondingly.

|          | Training Data | | Testing Data | |
|----------|:---:|:---:|:---:|:---:|
|          | $L_g$ | $L_r$ | $L_g$ | $L_r$ |
| Hybrid   | 0.690 | 1.228 | **0.869** | **1.409** |
| Global   | **0.640** | – | 0.910 | – |
| Pairwise | – | **1.197** | – | 1.420 |

Table 2. Average cross entropy for different training schemes.

#### 5.1.1 Global metric

We define an image as an attractive action shot if more than $p_a$ percentage of Turkers rate it as attractive (3 stars). For each specific $p_a$, we can draw a ROC curve as our evaluation metric for measuring the performance of our predicted attractiveness scores compared with the global ratings by Turkers. Once fixed $p_a$, like a binary classifier, we could slide the score threshold to discriminate if an image is positive or negative according to its attractiveness score predicted by our model. In our experiments, we fix $p_a$ as $0.2$ for global evaluation. Similar performance trend and pattern were observed when setting $p_a$ to other values.

#### 5.1.2 Pairwise metric

We define image $I^1$ to be more attractive than image $I^2$ if $p_{more} - p_{less} > p_b$. $p_{more}$ stands for the percentage of Turkers who rated $I^1$ to be *much more attractive* or *slightly more attractive* while $p_{less}$ indicates the percentage of Turkers who rated *much less attractive* or *slightly less attractive*. Similar to $p_a$, $p_b$ is also used to determine the ground-truth when conducting the comparison between our model and Turkers' rating. For each specific $p_b$, we can use the classification accuracy as the evaluation metric for the pairwise ratings. We choose different $p_b$ ($0.3, 0.4, 0.5, 0.6$) in our experiments. Note that, we can't draw ROC curves in such settings.

### 5.2. Hybrid training v.s. global/pairwise training

In our hybrid model, we propose to adaptively combine both the global and pairwise supervision. However, as shown in Figure 3, we could easily remove either the supervision component to perform the comparison. To see the benefits of our hybrid model, we first directly compare the cross entropy loss (defined in Equation 4 and 7) as shown in Table 2. On the training data, the global case produces the lowest $L_g$ while the pairwise case produces the lowest $L_r$. As to the testing data, our hybrid model has the lowest $L_g$ and $L_r$. This is because using only one supervision tends to overfit, thus performs worse on testing data.

The cross entropy evaluation is straightforward but does not help us gain a good insight. To achieve better understanding, we use the metrics mentioned in the last subsec-tion to perform further evaluation. Under both the global and pairwise metrics, as shown in the left column of Figure 4, it is clear that our proposed hybrid model outperforms other models that trained by using only the global ratings or pairwise ratings. Specifically, when comparing the performances using pairwise metric, as the pairwise model would output a distribution on the five relative attractiveness labels, it is natural to tell whether the first image is more attractive than the second one or not. However, as the global model only outputs an attractive score, one must set another threshold $\tau$ so that two images are regarded as equally good if there score difference is less than $\tau$. We have tried many thresholds and chose the best one for comparison. As shown in the bottom left subplot of Figure 4, our hybrid model always outputs the best performance on different $p_b$.

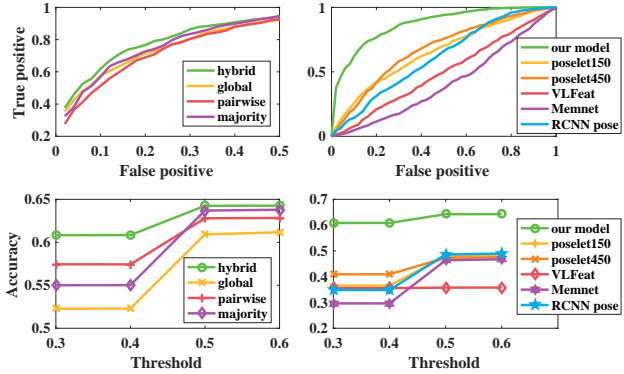

Figure 4. Performance comparisons under both global and pairwise metrics. The top row shows the comparison under global metric while the bottom row shows the comparison under pairwise metric. Note that our proposed model always outperforms the rest ones.

Moreover, we used the majority vote and learned a separate model with the consolidated unique ratings, under both metrics, it always performs worse than our hybrid model, this is perhaps because the simple majority vote from all Turkers discard some user preference information while our hybrid distribution loss would leverage all the crowdsourcing rating information, which further demonstrates the benefits of our proposed model.

### 5.3. Comparison to other methods

To further evaluate the performance of our prosed model, we extracted other features such as VLFeat [28], poselet [3], R-CNN pose [13] and memnet [21]. Although memorability attribute is intrinsically different from attractiveness, memnet is still the most recent and relevant high-level image attribute work among all other works, such as interestingness and aesthetics. For VLFeat, we extracted dense sift features and conducted dictionary learning and fisher vector coding. For poselet feature, we use the 150 categories as filters and use the highest activation as the value of the corresponding dimension. We use 1 or 3 pyramid levels. Thus the corresponding feature dimensions are 150 and 450. Denote the two kinds of methods as poselet150 and poselet450. We then trained a linear model on each type of feature using the same loss function as we proposed. As memnet outputs a memorability score for each image, so we can directly use that score for the comparison. As expected, our model outperforms all these methods under both global metric and pairwise metric, as shown in the right column of Figure 4. It is obvious that the ROC curve of our model is always above those of other methods. The mid level poselet feature is better than the low level VLFeat. The R-CNN pose feature is even worse than the poselet feature. It is because the feature dimension is very small and it doesn't explicitly define different modes of poses. The memnet is the worst because the score it outputs is designed for a different attribute, which means the most memorable images are not necessarily the most attractive ones. As demonstrated in Figure 6, although some of memorable images are indeed attractive (dance poses) but not all of them are attractive action shots.

### 5.4. Visualizations

What has our model learned from our dataset? We order all the 2000 test images from low score to high score and shown the images at different ranks in Figure 5(a). We select an image every 100 images. So the ranks of these images are $1, 101, 201, \ldots, 1901$. The score of each image is also listed in the figure. It is quite clear that the photos have high scores are more attractive. Such attractiveness order is not strict. For example, the first and third photos in the third row should have roughly the same rank. Though the order is not that accurate, it is roughly correct. The goal keeper picking up the ball in the sky (the last photo) is apparently much more attractive than most of the other images. The first photo is almost absolutely still.

We also visualize the image patches which have the highest activation and the lowest activation on the score neurons, as shown in Figure 5(b)(c). The neurons which have the highest activation are mostly the body parts with attractive poses for specific human actions in specific backgrounds, while the lowest activation neurons are primarily the standing straight body poses. The visualization is consistent with our hypothesis that the attractiveness of human action shot is jointly determined by both the human body pose and the background.

### 5.5. Applications

Our model could rate each action shot how attractive it is. We can apply it to some sports video clips like gymnastics, parkour, skate boarding *etc*. We randomly selected 80 clips sampled from [32] and ran our model to get the attractive scores for each frame. A score normalization ($[0, 1]$) is then performed within each clip. As shown in Figure 7, we can see in the first clip, our model produces a higher score when the person is jumping in the sky; while in the second clip, even through the skateboarder is partially occluded, our model is still able to output the most attractive frame within the sequence. In contrast, the score produced by the memnet [21] seems not reasonable. This is again showing that the memorability is not necessarily correlated with attractiveness for human action shots. To better understand how the attractive scores correlate with those peak action shots, we further asked judges to annotate the peak action shots for each of those sampled clips, and then we computed their average scores among all the clips. Their normalized average attractive score is $0.65$ for all the peak action shots.

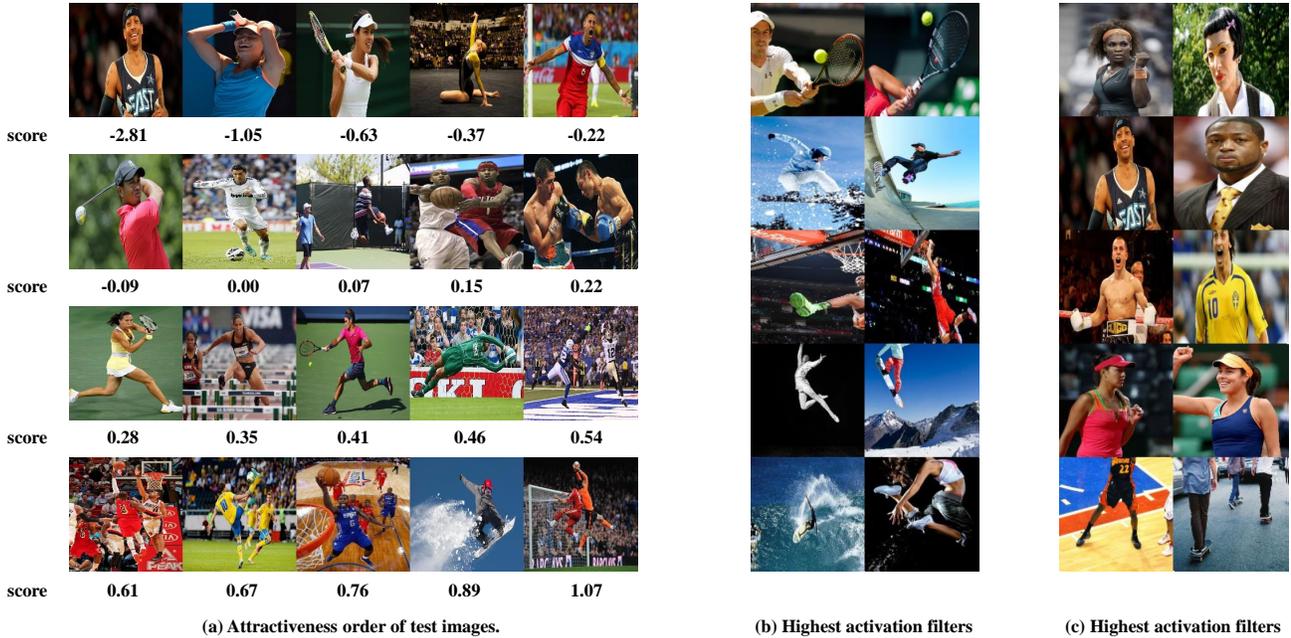

Figure 5. Visualization of image order and highest/lowest activation filters.

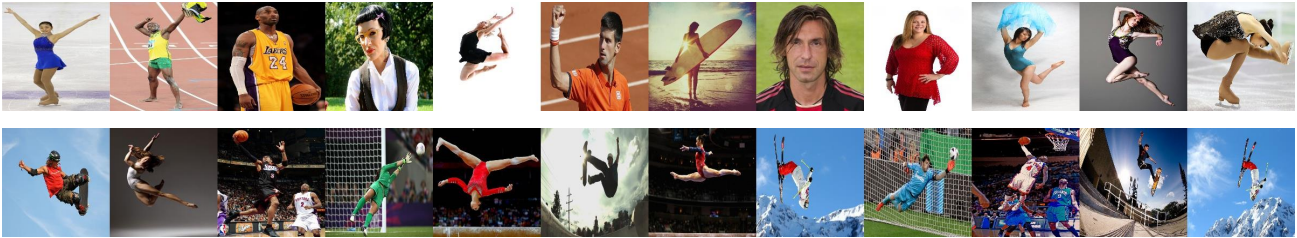

Figure 6. Most memorable (top) and attractive action shots (bottom) from test set

## 6. Conclusions and Limitations

In this paper, we introduced a new problem of predicating the attractiveness of human action shots. We collected about $8000$ human action shots from Internet and conducted rich crowd-scouring to annotate the degree of attractiveness in terms of both global and relative ratings. We then proposed a novel hybrid distribution matching loss function on top of a Siamese deep network structure to seamlessly integrate both types of ratings. Experiments showed that although subjective, the attractiveness attribute is predictable by our proposed model. However, as our current data was collected primarily targeting for studying the factors of human body pose and the surrounding context, we can see that people are always the most salient region in those action shots. Our model does not work well when the people are extremely small in the image. Moreover, thoroughly understanding the correlations between attractive action shots with other high-level attributes might be another interesting future work. Nevertheless, our work can still enable many interesting applications such as attractive action shot selection from a burst set or personal photo album.

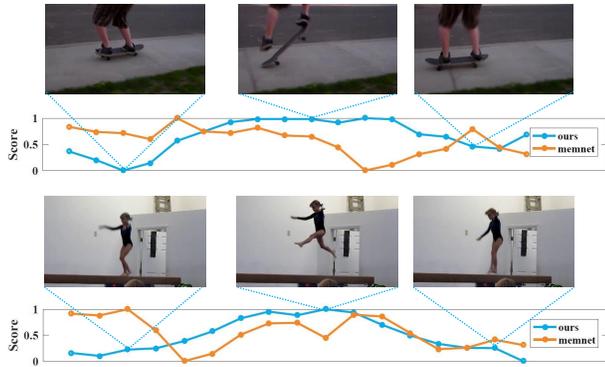

Figure 7. Score curves for two different action shot sequences. The blue curve was generated by our attractive model while the orange curve was generated by the memorability model (memnet [21]). Note that, the memorability values trend to be flat for these two sequences, the most memorable shot does not correspond to the most attractive peak action shot.

## References

[1] M. Andriluka, L. Pishchulin, P. Gehler, and B. Schiele. 2d human pose estimation: New benchmark and state of the art analysis. In *IEEE Conference on Computer Vision and Pattern Recognition (CVPR)*, June 2014.

[2] A. C. Berg, T. L. Berg, H. Daum, J. Dodge, A. Goyal, X. Han, A. Mensch, M. Mitchell, A. Sood, K. Stratos, and K. Yamaguchi. Understanding and predicting importance in images. In *Computer Vision and Pattern Recognition (CVPR), 2012 IEEE Conference on*, pages 3562–3569, June 2012.

[3] L. Bourdev and J. Malik. Poselets: Body part detectors trained using 3d human pose annotations. In *Computer Vision, 2009 IEEE 12th International Conference on*, pages 1365–1372. IEEE, 2009.

[4] A. Ceroni, V. Solachidis, C. Niederée, O. Papadopoulou, N. Kanhabua, and V. Mezaris. To keep or not to keep: An expectation-oriented photo selection method for personal photo collections. In *Proceedings of the 5th ACM on International Conference on Multimedia Retrieval*, ICMR '15, 2015.

[5] W.-T. Chu and C.-H. Lin. Automatic selection of representative photo and smart thumbnailing using near-duplicate detection. In *Proceedings of the 16th ACM International Conference on Multimedia*, MM '08, pages 829–832, 2008.

[6] R. Datta, D. Joshi, J. Li, and J. Z. Wang. Studying aesthetics in photographic images using a computational approach. In *Proceedings of the 9th European Conference on Computer Vision - Volume Part III*, ECCV'06, pages 288–301, 2006.

[7] S. Dhar, V. Ordonez, and T. L. Berg. High level describable attributes for predicting aesthetics and interestingness. In *Computer Vision and Pattern Recognition (CVPR), 2011 IEEE Conference on*, pages 1657–1664, June 2011.

[8] S. Dhar, V. Ordonez, and T. L. Berg. High level describable attributes for predicting aesthetics and interestingness. In *Proceedings of the 2011 IEEE Conference on Computer Vision and Pattern Recognition*, CVPR '11, pages 1657–1664, 2011.

[9] P. Dollar, C. Wojek, B. Schiele, and P. Perona. Pedestrian detection: An evaluation of the state of the art. *IEEE Transactions on Pattern Analysis and Machine Intelligence*, 34(4):743–761, April 2012.

[10] R. Dubey, J. Peterson, A. Khosla, M.-H. Yang, and B. Ghanem. What makes an object memorable? In *The IEEE International Conference on Computer Vision (ICCV)*, December 2015.

[11] Y. Fu, T. M. Hospedales, T. Xiang, S. Gong, and Y. Yao. Interestingness prediction by robust learning to rank, 2014.

[12] G. Gkioxari, R. Girshick, and J. Malik. Contextual action recognition with R*CNN. In *Proceedings of the International Conference on Computer Vision (ICCV)*, 2015.

[13] G. Gkioxari, B. Hariharan, R. Girshick, and J. Malik. R-cnns for pose estimation and action detection. *arXiv preprint arXiv:1406.5212*, 2014.

[14] M. Gygli, H. Grabner, H. Riemenschneider, F. Nater, and L. Van Gool. The interestingness of images. *ICCV*, 2013.

[15] P. Isola, D. Parikh, A. Torralba, and A. Oliva. Understanding the intrinsic memorability of images. In *Advances in Neural Information Processing Systems*, 2011.

[16] P. Isola, J. Xiao, A. Torralba, and A. Oliva. What makes an image memorable? In *IEEE Conference on Computer Vision and Pattern Recognition (CVPR)*, pages 145–152, 2011.

[17] M. Jas and D. Parikh. Image specificity. In *The IEEE Conference on Computer Vision and Pattern Recognition (CVPR)*, June 2015.

[18] Y.-G. Jiang, Y. Wang, R. Feng, X. Xue, Y. Zheng, and H. Yang. Understanding and predicting interestingness of videos. In *AAAI*, 2013.

[19] S. E. Kahou, X. Bouthillier, P. Lamblin, Ç. Gülçehre, V. Michalski, K. R. Konda, S. Jean, P. Froumenty, Y. Dauphin, N. Boulanger-Lewandowski, R. C. Ferrari, M. Mirza, D. Warde-Farley, A. C. Courville, P. Vincent, R. Memisevic, C. J. Pal, and Y. Bengio. Emonets: Multimodal deep learning approaches for emotion recognition in video. *CoRR*, abs/1503.01800, 2015.

[20] A. Khosla, A. Das Sarma, and R. Hamid. What makes an image popular? In *Proceedings of the 23rd International Conference on World Wide Web*, WWW '14, pages 867–876, 2014.

[21] A. Khosla, A. S. Raju, A. Torralba, and A. Oliva. Understanding and predicting image memorability at a large scale. In *International Conference on Computer Vision (ICCV)*, 2015.

[22] X. Lu, Z. Lin, H. Jin, J. Yang, and J. Z. Wang. Rapid: Rating pictorial aesthetics using deep learning. In *Proceedings of the 22Nd ACM International Conference on Multimedia*, MM '14, 2014.

[23] X. Lu, Z. Lin, X. Shen, R. Mech, and J. Z. Wang. Deep multi-patch aggregation network for image style, aesthetics, and quality estimation. In *Proceedings of the IEEE International Conference on Computer Vision*, pages 990–998, 2015.

[24] S. Ren, X. Cao, Y. Wei, and J. Sun. Face alignment at 3000 fps via regressing local binary features. In *The IEEE Conference on Computer Vision and Pattern Recognition (CVPR)*, June 2014.

[25] F. Schroff, D. Kalenichenko, and J. Philbin. Facenet: A unified embedding for face recognition and clustering. In *Computer Vision and Pattern Recognition (CVPR), 2015 IEEE Conference on*, pages 815–823, June 2015.

[26] K. Simonyan and A. Zisserman. Very deep convolutional networks for large-scale image recognition. *arXiv preprint arXiv:1409.1556*, 2014.

[27] P. Sinha, S. Mehrotra, and R. Jain. Summarization of personal photologs using multidimensional content and context. In *Proceedings of the 1st ACM International Conference on Multimedia Retrieval*, ICMR '11, 2011.

[28] A. Vedaldi and B. Fulkerson. VLFeat: An open and portable library of computer vision algorithms, 2008.

[29] R. L. Vieriu, S. Tulyakov, S. Semeniuta, E. Sangineto, and N. Sebe. Facial expression recognition under a wide range of head poses. In *Automatic Face and Gesture Recognition (FG), 2015 11th IEEE International Conference and Workshops on*, volume 1, pages 1–7, May 2015.

[30] P. Viola and M. Jones. Robust real-time object detection. In *International Journal of Computer Vision*, 2001.

[31] T. C. Walber, A. Scherp, and S. Staab. Smart photo selection: Interpret gaze as personal interest. In *Proceedings of the SIGCHI Conference on Human Factors in Computing Systems*, CHI '14, pages 2065–2074, 2014.

[32] H. Yang, B. Wang, S. Lin, D. Wipf, M. Guo, and B. Guo. Unsupervised extraction of video highlights via robust recurrent auto-encoders. In *IEEE International Conference on Computer Vision. ICCV'15, Santiago, Chile*, Dec 2015.

[33] J.-Y. Zhu, A. Agarwala, A. A. Efros, E. Shechtman, and J. Wang. Mirror mirror: Crowdsourcing better portraits. *ACM Transactions on Graphics (SIGGRAPH Asia 2014)*, 33(6), 2014.